\documentclass{article}
\usepackage{ijcai21}

\usepackage{times}
\usepackage{soul}
\usepackage{url}
\usepackage[hidelinks]{hyperref}
\usepackage[utf8]{inputenc}
\usepackage[small]{caption}
\usepackage{graphicx}
\usepackage{amsmath}
\usepackage{amsthm}
\usepackage{booktabs}
\usepackage{algorithm}
\usepackage{algorithmic}
\usepackage{caption}
\usepackage{lipsum}
\usepackage[english]{babel}
\newtheorem{theorem}{Theorem}[section]
\newtheorem{corollary}{Corollary}[theorem]
\newtheorem{lemma}[theorem]{Lemma}
\DeclareMathOperator*{\argmin}{argmin} 

\title{On reducing the order of arm-passes bandit streaming algorithms under memory bottleneck}

\usepackage{xcolor}

\author{
Santanu Rathod$^1$
\affiliations
$^1$IIT-Bombay\\
\emails
santanusrathod@gmail.com
}

\begin{document}

\maketitle

\begin{abstract}
  In this work we explore multi-arm bandit streaming model, especially in cases where the model faces resource bottleneck. We build over existing algorithms conditioned by limited arm memory at any instance of time. Specifically, we improve the amount of streaming passes it takes for a bandit algorithm to incur a $O(\sqrt{T\log(T)})$ regret by a logarithmic factor, and also provide 2-pass algorithms with some initial conditions to incur a similar order of regret.  
\end{abstract}

\section{Introduction}
In this paper we improve upon previous work done on regret minimization in Multi-Armed Bandit (MAB) \cite{Berry et al.} in constrained memory setting. In the multi-armed bandit setting, in general, the arms can be thought as handles in slot machines, with each handle giving a reward each time we pull it; the rewards for a particular arm follow an instance dependent but fixed distribution. The objective is to maximize the rewards accumulated in the end, or to put it in another way learn a policy to pull arms in a particular(regret minimizing or reward maximizing) manner. 
\\\\
With multi-arm bandits having abundant applications in healthcare, finance, dynamic-pricing models, recommender systems, etc. it makes sense to think about budgetary and resource constraints, especially in settings where the number of levers are large. In \cite{Shen et al.}, the authors proposed a bandit algorithm for making online portfolio choices via exploiting correlations among multiple arms, now given that high frequency trading firms operate on large volumes of data \textit{frequently} they can surely benefit from near optimal performance in resource constraint setting. Even in social good problems like drug-testing \cite{Armitage et al.}, and other practically relevant tasks like crowd-sourcing \cite{Tran et al.} that are modelled using multi armed bandits it makes economic sense to explore options that operate near optimally without storing entire statistics of the system \\\\
To that end, in \cite{Chaudhari et al.} the authors have established an instance independent $O(\sqrt{Tlog(T)})$ bound with $O(log(T))$ streaming passes and recently \cite{Maiti et al.} have established a lower bound of $O(T^{2/3})$ for any single pass algorithm. Our objective thus was to explore the domain between $log(T)$ and one pass, and try to minimise the regret accumulated while also reducing the number of passes to the extent that we can.\\\\
Our contributions:
\begin{itemize}
    \item We propose a variation of Algorithm-1 in \cite{Chaudhari et al.} which accrues $O(\sqrt{Tlog(T)})$ regret with $log(log(T))$ passes instead of $log(T)$ passes.
    \item Using the analysis devised for above result we then propose a 2-pass-algorithm, with some instance dependent initial conditions, with $O(\sqrt{Tlog(T)})$ regret.
    \item Simulations to corroborate the results.
\end{itemize}
We also include an instance dependent two-pass-hybrid algorithm with some prior information about the system which incurs a $log(T)$ regret. In the following sections we first start by describing the streaming model, RAM model, that we use, describe the algorithms and the key intuitions behind proving the result, and then doing same with 2-pass algorithms. 
\section{Related work}
Right from the seminal work of \cite{Robbins et al.} the predominant body of literature in stochastic multi-armed bandit is dedicated to the regret minimisation task on finite and infinite bandit instances. Later, a number of salient algorithms like UCB1 (\cite{Auer et al.}), Thompson Sampling (\cite{Chapelle et al.}; \cite{Agarwal et al.}),  have been shown to achieve the order optimal cumulative regret on the finite instances. The study on multi arm bandit algorithms under constraint resources however is still limited despite its myriad practical applications today. \cite{Liau et al.} where they provide an instance dependant optimal regret with O(1) storage of arms and \cite{Chaudhari et al.} providing instance independent $\sqrt{Tlog(T)}$ regret with $log(T)$ passes. These results can however be further improved as shown in the following sections.
\\\\
Since early 60s and 50s finite memory hypothesis testing has been looked at by researchers (\cite{Robbins et al.}; \cite{Cover76 et al.}). In multi armed bandit setting \cite{Cover68 et al.} first presented a finite memory algorithm for two-armed Bernoulli instance, achieving an average reward which converges to optimality, with high probability. The approach consisted of a collection of interleaved test and trial blocks, where each test block is divided into several sub-blocks and the switching among these sub-blocks is governed by a finite state ma- chine. However, he considered only two-armed Bernoulli 
instances, and the approach guarantees only an asymptotic convergence of the empirical average reward. Hence, this setup is not very interesting, as our objective is to present a finite-time analysis of regret for general bandit instances. \\
To that end, we now present our bandit streaming setup and corresponding algorithms and analysis.
\\\\
\section{Preliminaries}
\subsection{RAM-Model}
It should be noted that given any bandit instance $B = (A, D)$, as we are not considering any special structure in $A,D$, putting a restriction on an algorithm to use a bounded number of words of space, either restricts the horizon of pulls, or restricts the algorithm to store statistics of only bounded number of arms simultaneously. 
In this paper, we consider the latter and assume $M$ to be that number. We adopt the word RAM model (\cite{Aho et al.}, \cite{Cormen et al.}), that considers a word as the unit of space. This model facilitates to consider that each of the in- put values and variables can be stored in $O(1)$ word space. For finite bandit instances $(|A|<\infty)$, we consider a word to be consisted of $O(log(T))$ bits. Therefore, our algorithm needs space-complexity of $O(Mlog(T)+log(|A|))$ bits.
We call this set of arm indices whose statistics are stored as arm memory and its cardinality as arm memory size. Hence, an algorithm with arm memory size M can store the statistics of at most M arms. Also, it should be noted that an algorithm is allowed to pull an arm only if it is stored in the memory. Hence, before pulling a new arm (which is not currently in the arm memory), the algorithm should replace an arm in its arm memory with this new arm. It is interesting to note that the algorithms that work with M = 1, can only keep the stat of the arm it is currently pulling. There- fore, switching to a new arm costs such an algorithm to lose all the experience gained by sampling the previous arm. However, for a finite bandit instance, as the algorithms are allowed to remember all the arm indices, such an algorithm can store the gained experience by storing a bounded number of arm indices for possible further special treatment.

\subsection{Simple Regret}
It is one of the popular problems in \textit{pure exploration} bandit setting which focuses on the design of strategies making the best possible use of available numerical resources (e.g., as cpu time) in order to optimize the performance of some decision-making task. If $b_{t} \in A$ is the arm recommended by the algorithm after $t-th$
pull, then the simple regret at $t$ is defined as, 
\begin{equation}
    E[r^{*}_{t}]= \mu^{*}- E[\mu_{b_{t}}]
\end{equation}
\\\\
In particular \cite{Bubeck et al.} studies the relationship between simple regret and cumulative regret, with results showing that upper bounds on cumulative regret should also lead to upper bounds on simple regret. \\\\
Figure 1. from \cite{Bubeck et al.} shows the framework of a pure-exploration problem. Where the forecaster can be interpreted as the algorithm in consideration for the bandits, e.g. UCB1 or Thompson sampling. The forecaster then, at the end of its duration, recommends a particular arm based on some criteria, which can for example be a. Most played arm, b. Empirically best arm, or c. Arm with highest upper confidence and so forth. In this study we're concerned with most played arm(MPA) as our recommendation strategy. 
\begin{theorem}[Distribution-free upper bound on Simple-Regret of UCB-MPA by \cite{Bubeck et al.}] \textit{Given a K-
sized set of arms $K$ as input, if UCB-MPA runs for a horizon of $T$ pulls such that $T \geq K(K + 2)$, then for some
constant $C > 0$, it achieves the expected simple regret
} $E[r^{*}_{T}] \leq C\sqrt{\dfrac{K\log(T)}{T}}$. 
\end{theorem}
\begin{figure}[H]
\centering
\includegraphics[width= 1\linewidth]{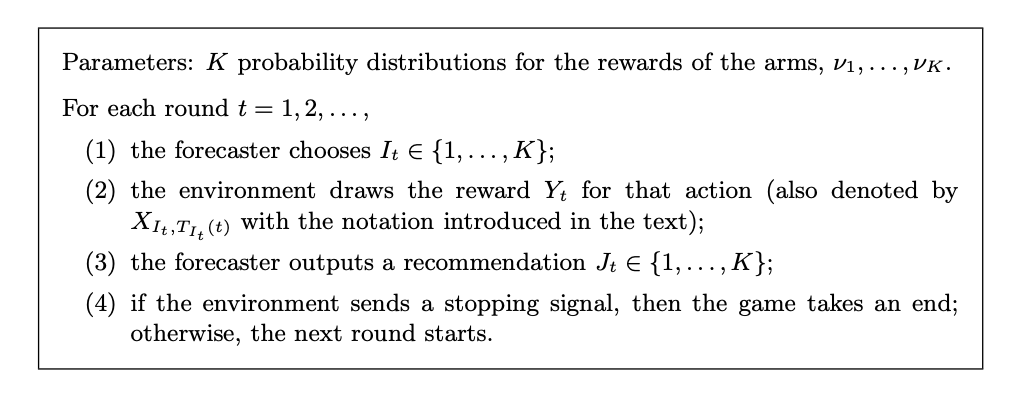}
\caption{The pure exploration problem for multi-armed bandits (with a finite number of arms)}
\end{figure}

\textbf{From hereon \textit{ln}: logarithm with base $e$ and $\log$: logarithm with base 2.}
\section{Multi pass algorithm}
Algorithm 1, which we call UCB-LAM (UCB-limited arm memory), is presented in this section. We establish the upper-bound on cumulative regret and also the improvement in the order of passes over UCB-M proposed in \cite{Chaudhari et al.}. The allocation strategy in UCB-LAM is UCB1 \cite{Auer et al.} for now, although it can in principle be MOSS, Thompson Sampling, etc. After exploring this multi pass algorithm we then move on to studying constant pass algorithms.

\subsection{Intuition behind UCB-LAM} 
We know that main driving idea behind any multi-arm bandit algorithm focused on minimising regret is to sample rewards from the best arm with time. Although when faced with resource bottlenecks it's not even guaranteed that the best arm will be present in the arm memory, let alone consistently pulling it. Thus, intuitively, any algorithm that's minimizing regret under limited arm memory ought to take care that, a. The probability of the best arm being present in arm memory increases with time, b. Given (a) the best arm is sampled often. To that end we work towards UCB-LAM and also provide probabilistic analysis solidifying our intuitions, (a) and (b).

\subsection{Description of UCB-LAM}
In algorithm 1, UCB-LAM, we're given $A:$ a set of $K$ arms, $K > M \geq 2:$ arm memory size. Now when $M \geq K$ we can simply run UCB1 \cite{Auer et al.} on the arms $A$ since the arm memory can accommodate all the arms. But when $M < K$ we sample arms in phases. There are a total of $x_{0}$ phases and each phase $w$ is divided into $h_{0} = \lceil \dfrac{K-1}{M-1} \rceil$ sub-phases to accommodate appropriate $M-$sized subset of arms $S^{w, j}$ that're allowed in the memory for phase $w$, and sub-phase $j$. Once the Allocation Strategy is applied on $S^{w,j}$ the sub-phase then recommends $\hat{a} \in  S^{w,j}$ to the next sub-phase $S^{w, j+1} \text{or}  S^{w+1, 1}$. \\\\
Each of the $h_{0}$ sub-phases present in phase $w$ are allotted a time-duration of $b_{w}$, with $b_{1}=M(M+2)$ \{change? Why?\} and with the time allotted to a sub-phase increasing as the phase increases with $b_{w}= b_{w-1}^{2}$. The increasing time allocation with phase indicates the algorithms growing confidence in the best arm being present in arm memory and it being sampled often.

 

\subsection{Preliminary arguments for regret analysis}
We note that with $K$ arms, $M$ memory size, and horizon $T$, the number if sub-phases is $h_{0}= \lceil \frac{K-1}{M-1} \rceil$ and the total number of phases or arm-passes is upper bounded by $x_{0}= 1+ \lceil \log(\log_{M(M+2)}( \dfrac{T}{h_{0}} )) \rceil$ (Lemma A.1 in Appendix A).
\\\\
As we'll further see in our regret bifurcation and analysis further ahead, we need to upper-bound the mean of the recommended arms between two consecutive sub-phases. Let $a^{y,j}_{*} \in S^{y,j}$ 
be the arm recommended by sub-phase $j$ to $j+1$, then we need to bound $r^{y, j}= \mu_{*}^{y,j-1}- \mu_{*}^{y, j}$. Since we know that, $\mu_{*}^{y, j-1} \leq \max_{a \in S^{y,j}} \mu_{a}$, we can write \[r^{y, j}= \mu_{*}^{y,j-1}- \mu_{*}^{y, j} = \max_{a \in S^{y,j}} \mu_{a}- \mu_{*}^{y, j} - \epsilon \]
We thus get:
\begin{corollary}
\textit{Using Theorem 3.1 in phase} $y$, \textit{at the end of each sub-phase} $j$, \textit{the approximate simple regret with respect to} $\mu_{a_{*}^{y,j}}$ \textit{is upper bounded as} $E[r^{y,j}] \leq C\sqrt{\dfrac{K\log(b_{y})}{b_{y}}}$.
\end{corollary}
We also need to be cognizant of the fact that the arm recommended by a sub-phase might not be the optimal arm, because in the worst case scenario the allocation strategy will only be sampling sub-optimal arms and thus incurring a huge regret. It is thus imperative to bound the \textit{approximate} simple regret and to take care that optimal arm is included the arm-memory, and once included, it is also the arm that's recommended to the next sub-phase with high probability. We'll come across these arguments naturally while doing the regret analysis.
\begin{lemma}
Consider events, 
\begin{enumerate}

    \item A: Best arm(universal) is in the current ($w,j$) instance.
    \item B: Best arm(universal) is recommended to next instance (($w, j+1$) or ($w+1, 1$)).
    
\end{enumerate}
Let $s=(w,j)$ be the current instance with phase $w$ and sub-phase $j$, and let $r$ be the arm recommended to the next instance $(w, j+1) or (w+1, j)$ and let $\mu_{*}^{w,j}$ be the mean of that arm. Then if 
\begin{itemize}
    \item $T_{1}= P(A_{s}=1)\times \{b_{w}(E[\mu^{*}- \mu_{*}^{w,j}])P(B_{s}=0|A_{s}=1)$
    \item $T_{2}= P(A_{s}=0)\times \{b_{w}(E[\mu^{*}- \mu_{*}^{w,j}]) \}$
\end{itemize}
We have, \\
$\dfrac{T_{1}+T_{2}}{b_{w}} \leq 2C\dfrac{(M-1)\cdot h_{0}}{(b_{w-1}/M -1)^{2}}(h_{0}+1)\sqrt{\dfrac{M\log(b_{w-1})}{b_{w-1}}}$ \\

\end{lemma}
The proof is presented in the Appendix A. Next we birfurcate the regret appropriately and use the above results to find the upper bound the cumulative regret ($R_{T}^{*}$)
\\\\
\textbf{Bifurcation of $R^{*}_{T}$}. For any given phase $w$, and a sub-phase $j$, let $\mu_{*}^{w,j}=$ mean of the most played arm $a$ for $a\in S^{w,j}$, and $R_{w,j}$ be the incurred regret. Then, 

\begin{align*}
R_{w,j} &= b_{w}\mu^{*} - \sum_{t=1}^{b_{w}} E[\mu_{a_{t}}] \\ 
&= b_{w}E[\mu^{*}- \mu_{*}^{w,j}] + \sum_{t=1}^{b_{w}}(E[\mu_{*}^{w,j}] -E[\mu_{a_{t}}])
\end{align*}

Now let $R^{(1)}_{w,j}= b_{w}(\mu^{*}- \mu_{*}^{w,j})$, and $R^{(2)}_{w,j} = \sum_{t=1}^{b_{w}} (E[\mu_{*}^{w,j}] - E[\mu_{a_{t}}])$, we write,

\begin{equation}
R^{*}_{T}= \sum_{w=1}^{x_{0}}\sum_{j=1}^{h_{0}}R_{w,j}= \sum_{w=1}^{x_{0}}\sum_{j=1}^{h_{0}} (R^{(1)}_{w,j} + R^{(2)}_{w,j}) 
\end{equation}

Having bifurcated cumulative regret $R^{*}_{T}$ in terms of $R^{1}, R^{2}$ we can see that minimizing $R^{1}$ would essentially mean that with high probability, as we go further in phases, the best arm in memory is actually the optimal arm. As can be seen from \textbf{Lemma A.2} for phase $w$, the P(best arm is recommended)= $P(J_{b_{w}}= i^{*}) \geq 1- \dfrac{K-1}{\alpha -1}(b_{w}/K -1)^{2(\alpha-1)}$, for $\alpha=2$, increases with phase, which is also the key argument that we use while proving \textbf{Lemma 4.2} below. \\\\

\begin{lemma}
 For $2 \leq M < K$, and for $T \geq KM^{2}(M+2)$, let $R^{(1)}= \sum_{w=1}^{x_{0}}\sum_{j=1}^{h_{0}} R^{(1)}_{w,j}$, then, 
\[ 
R^{1} \leq C_{0}+ C_{2} \times \log(\log_{b_{1}}(\dfrac{T}{h_{0}}))
\]
where $C_{2}$ is a constant depending on $K,M$.
\end{lemma}
For the proof, refer Appendix A. For calculating $\sum_{w=1}^{x_{0}}\sum_{j=1}^{h_{0}} R^{(2)}_{w,j}$, can be upper-bounded using the
problem independent upper-bound on the cumulative regret
of UCB1 \cite{Auer et al.}, which we restate below. \\\\

And $R^{2}$ can be interpreted as a local cumulative regret, or the regret accrued with respect to the best arm present in a particular memory instance. In the worst case scenario we expect the arms to be present in increasing order of means, and thus the local regret then would thus increase linearly with instances and will be within a certain factor of the local cumulative of the instance containing best arm. This what \textbf{Lemma 4.4} puts mathematically. 
\begin{lemma}
(Distribution-Free Upper Bound on Cumula-
tive Regret of UCB1 \cite{Auer et al.}). Given a set of K-
arms as the input, for any horizon
T, the cumulative regret incurred by UCB1 $R^{*}_{T} \leq 12\sqrt{TK\log(T)}+6K$. Further, if $T \geq K/2$, then $R^{*}_{T} \leq 18\sqrt{TK\log(T)}$.
\end{lemma}
Next using \textit{Lemma 4.3} we upper bound  $\sum_{w=1}^{x_{0}}\sum_{j=1}^{h_{0}} R^{(2)}_{w,j}$, proof is given in Appendix A. 
\\\\
\begin{lemma}
For $2 \leq M < K$, and for $T \geq KM^{2}(M+2)$, let,  $R^{(2)}= \sum_{w=1}^{x_{0}}\sum_{j=1}^{h_{0}} R^{(2)}_{w,j}$, then,  
\[
R^{(2)} \leq C_{0}+ C_{4}\sqrt{(\log_{b_{1}}(\dfrac{T}{h_{0}}))\cdot(\dfrac{T}{h_{0}})}
\]
Where $C_{0} , C_{4}$ are constants depending on K and M.
\end{lemma}

\begin{theorem}[Main result]
For $2 \leq M < K$, and for $T \geq KM^{2}(M+2)$, let $R^{(1)}= \sum_{w=1}^{x_{0}}\sum_{j=1}^{h_{0}} R^{(1)}_{w,j}$ and $R^{(2)}= \sum_{w=1}^{x_{0}}\sum_{j=1}^{h_{0}} R^{(2)}_{w,j}$, so for cumulative regret $R^{*}_{T}= R^{(1)}+R^{(2)}$ we have, 
\[
R^{*}_{T} \leq 2C_{0}+C_{2}\times (\log(\log_{b_{1}}(\dfrac{T}{h_{0}})))+C_{4}\sqrt{(\log_{b_{1}}(\dfrac{T}{h_{0}}))\cdot \dfrac{T}{h_{0}}}
\]
where $C_{0}, C_{1}, C_{4}$ are functions of $K,M$.
\end{theorem}
\begin{proof}
 One can easily see how combining bounds from \textbf{Lemma 4.2} and \textbf{Lemma 4.4} will give us the required upper-bound on cumulative regret of \textbf{Algorithm 1}.
\end{proof}

\begin{algorithm}[tb]
\caption{UCB-LAM(limited arm memory)}
\label{alg:algorithm_ucblam}
\textbf{Input}: $A:$ the set of $K$ arms indexed by [K], $M(\geq 2)$: Arm memory size\\

\begin{algorithmic}[1] 

\IF {$M \geq K$}
\STATE Run UCB1 on $A$ until horizon
\ELSE
\STATE $b_{1}= M(M+2).$ \{Initial horizon per sub-phase\}
\STATE $\hat{a}= 1$. \{ Initial arm recommendation\}
\STATE $w= 1$. \{Counts the number of phases \}
\STATE $h_{0}= \lceil \dfrac{K-1}{M-1} \rceil$. \{ The number of sub-phases in a phase\}
\STATE No random shuffling for now
\WHILE{ the horizon is not finished}
\STATE $l= 0$
\FOR {$j= 1, \cdots, h_{0};$ if the horizon is not finished}
\STATE $S^{w,j}= \{l+1, \cdots, \min\{l+1+(M-1), K\} \} $
\STATE $l=$ The highest arm index in $S^{w,j}$.
\IF { $\hat{a} \notin S^{w,j}$ }
\STATE $S^{w,j} = \{\hat{a}\} \cup S^{w,j}\setminus \{l\}$.
\STATE $l= l-1$
\ENDIF
\STATE \{ALLOCATION STRATEGY\}
\STATE Run UCB1 on $S_{w,j}$ for horizon of $b_{w}$ pulls
or the remaining horizon; whichever is smaller.
\STATE \{RECOMMENDATION STRATEGY\}
\STATE $\hat{a}=$ The most played arm in $S^{w,j}$
\ENDFOR
\STATE $w= w+1$. \{Increment phase count\}
\STATE $b_{w}= (b_{w-1})^{2}$. \{ Increment horizon per sub-phase\}
\ENDWHILE
\ENDIF


\end{algorithmic}
\end{algorithm}

\section{Constant pass algorithms}
Now that we've introduced \textbf{UCB-LAM} which gives us $O(\sqrt{T\log(T)})$ cumulative regret in $O(\log(\log(T)))$ passes, we further explore constant pass algorithms in similar spirit. Decreasing the number of passes is practically very desirable since it frees up the memory reservoir required to store the dormant arms not being considered by the allocation strategy at any instance\{come up with better motivation\}, and in extension having \textit{constant} number of passes will aid in determining when a section of memory space will be freed (after $c$ passes) and thereby enabling us to better plan the use of the memory as opposed to the case where the number of passes is dependant on the time horizon thereby restricting the scope of any prior planning for the memory in consideration.

\subsection{2-pass UCB-LAM}
2-pass UCB-LAM is an extension of UCB-LAM when restricted to two passes. The with the only difference being for us to know the time horizon beforehand -- which is obvious. Since when we say that a certain algorithm takes 2-passes, we ought to know \textit{when the first pass ends} and \textit{when the second pass begins}. \\\\

\begin{algorithm}[ht]
\caption{2-pass UCB-LAM}
\label{alg:algorithm_2passucblam}
\textbf{Input}: $A:$ the set of $K$ arms indexed by [K], $M(\geq 2)$: Arm memory size, $T:$ Total time horizon, large enough such that $T \geq K*(1+\dfrac{4\alpha ln(T)}{M \Delta_{i}^{2}})$ \\

\begin{algorithmic}[1] 

\IF {$M \geq K$}
\STATE Run UCB1 on $A$ until horizon
\ELSE
\STATE $\hat{a}= 1$. \{ Initial arm recommendation\}
\STATE $w= 1$. \{Counts the number of phases \}
\STATE $h_{0}= \lceil \dfrac{K-1}{M-1} \rceil$. \{ The number of sub-phases in a phase\}
\STATE $b_{1}= \dfrac{\sqrt{1+4\dfrac{T}{h_{0}}}-1}{2}$ 
\STATE $b_{2}= \dfrac{2\dfrac{T}{h_{0}}+1-\sqrt{1+\dfrac{4T}{h_{0}}}}{2}= b_{1}^{2}$ 
\WHILE{ the horizon is not finished}
\STATE $l= 0$
\FOR {$j= 1, \cdots, h_{0};$ if the horizon is not finished}
\STATE $S^{w,j}= \{l+1, \cdots, \min\{l+1+(M-1), K\} \} $
\STATE $l=$ The highest arm index in $S^{w,j}$.
\IF { $\hat{a} \notin S^{w,j}$ }
\STATE $S^{w,j} = \{\hat{a}\} \cup S^{w,j}\setminus \{l\}$.
\STATE $l= l-1$
\ENDIF
\STATE \{ALLOCATION STRATEGY\}
\STATE Run UCB1 on $S_{w,j}$ for horizon of $b_{w}$ pulls
or the remaining horizon; whichever is smaller.
\STATE \{RECOMMENDATION STRATEGY\}
\STATE $\hat{a}=$ The most played arm in $S^{w,j}$
\ENDFOR
\STATE $w= w+1$. \{Increment phase count\}
\ENDWHILE
\ENDIF
\end{algorithmic}
\end{algorithm}

From \textbf{Lemma B.1} we know for sure that the total number of passes required by \textbf{Algorithm-2} is two. Having said that we now put across the regret analysis and argument for \textbf{Algorithm-2}.
\subsubsection{Bifurcation of regret}
We know that the regret obtained from \textbf{Algorithm 2 $R^{*}_{T}$} can be bifurcated as, following steps as we did the regret bifurcation of \textbf{Algorithm-1}-

For any given phase $w$, and a sub-phase $j$, let $\mu_{*}^{w,j}=$ mean of the most played arm $a$ for $a\in S^{w,j}$, and $R_{w,j}$ be the incurred regret, Then, 

\begin{align*}
R_{w,j} &= b_{w}\mu^{*} - \sum_{t=1}^{b_{w}} E[\mu_{a_{t}}] \\
&= b_{w}E[\mu^{*}- \mu_{*}^{w,j}] + \sum_{t=1}^{b_{w}}(E[\mu_{*}^{w,j}] -E[\mu_{a_{t}}]).
\end{align*}

Now let $R^{(1)}_{w,j}= b_{w}(\mu^{*}- \mu_{*}^{w,j})$, and  \\ $R^{(2)}_{w,j} = \sum_{t=1}^{b_{w}} (E[\mu_{*}^{w,j}] - E[\mu_{a_{t}}])$, we write,

\[
R^{*}_{T}= \sum_{w=1}^{2}\sum_{j=1}^{h_{0}}R_{w,j}= \sum_{w=1}^{2}\sum_{j=1}^{h_{0}} (R^{(1)}_{w,j} + R^{(2)}_{w,j}) 
\]
We now deal with $R_{1}$ and $R_{2}$ separately as we did earlier.

\begin{lemma} For $2 \leq M < K$, and total time-horizon $T$, \\ we get that, for Algorithm-2,
\[
R_{2} \leq C_{2}+C_{0}\sqrt{T+0.25h_{0}}+C_{1}\sqrt{T\log(T/h_{0})}
\]
Where $C_{0}, C_{1}, C_{2}$ are constants depending on $M,K$.
\end{lemma}

As we can see the order of $R_{2}-regret$ accrued by \textbf{Algorithm-2} almost same as the order of $R_{2}-regret$ accrued by \textbf{Algorithm-1}. Essentially $R_{2}$ here can be interpreted as the summation of all the \textit{local} regrets, where local means being limited to one's memory instance without being aware of all the arms. We now move onto $R_{1}$.

\begin{lemma} For $2 \leq M < K$, and total time-horizon $T$, \\ we get that, for Algorithm-2,\\  $R_{1}= \sum_{w=1}^{2}\sum_{j=1}^{h_{0}}(R^{(1)}_{w,j}) \leq O(\sqrt{T})$
\end{lemma} 
\{refer Appendix for the proof\}. 
$R^{1}$ here can be physically interpreted as the penalty we accrue as the result of recommending sub-optimal arm instead of the best arm. While there can be several reasons for this, like a. Best arm is not present in the current memory instance, b. Best arm is present but isn't recommended for some reason, and all of these factors are analysed for in the proofs. \{comment on how $R_{1}$ here differs from $R_{1}$ for algorithm-1\}

\begin{theorem}
Given a set of K arms A with $K \leq 3$, an arm memory of size M, and the total time horizon T Algorithm-2 will incur a cumulative regret $R^{*}_{T}= O(A_{1}\sqrt{T}+A_{2}\sqrt{T\log(T)})$, where $A_{1}, A_{2}$ are constants depending on $K,M$.
\end{theorem}
\begin{proof}
For phase $w$, and sub-phase $j$, we know that, $R^{(1)}_{w,j}= b_{w}(\mu^{*}- \mu_{*}^{w,j})$, and $R^{(2)}_{w,j} = \sum_{t=1}^{b_{w}} (E[\mu_{*}^{w,j}] - E[\mu_{a_{t}}])$, and we write,

\[
R^{*}_{T}= \sum_{w=1}^{x_{0}}\sum_{j=1}^{h_{0}}R^{*}_{w,j}= \sum_{w=1}^{x_{0}}\sum_{j=1}^{h_{0}} (R^{(1)}_{w,j} + R^{(2)}_{w,j}) 
\]
And from \textbf{Lemma 5.1} we know that
\[
R_{2} \leq C_{2}+C_{0}\sqrt{T+0.25h_{0}}+C_{1}\sqrt{T\log(T/h_{0})}
\]
,and from \textbf{Lemma 5.2} we know that $R_{1} \leq O(\sqrt{T})$, thus we get, 
\begin{equation}
    R^{*}_{T}=R_{1}+R_{2} \leq O(A_{1}\sqrt{T}+A_{2}\sqrt{T\log(T)})
\end{equation}
where $A_{1}, A_{2}$ are constants depending on $K,M$.
\end{proof}
\subsection{Constant pass algorithm from a kind-of explore-exploit perspective}
As we've noted in the sections above, to accrue lesser regret in resource bottleneck settings intuitively what any algorithm tries to do is maximize the probability of the best arm being in the arm memory and being recommended often. In this subsection we explore a relatively simple algorithm that performs somewhat better than \textbf{2-pass UCB-LAM} theoretically speaking, building up on exactly the points we mentioned. \\
We'll first start with a \textbf{pseudo algorithm} and then based on our analysis build up \textbf{Algorithm-4}.
\begin{algorithm}\caption{(\textbf{pseudo}) 2-pass-hybrid}\label{alg:algorithm_2passhybrid}
\textbf{Input}: $A:$ the set of $K$ arms indexed by [K], $M(\geq 2)$: Arm memory size, $T:$ Total time horizon \\
\begin{algorithmic}[1]
\STATE \textbf{1st Pass:} 
\STATE   \text{Uniformly play all the arms (details later)}
\STATE   \text{Recommend the arm with highest} $\overline{\mu}$
\STATE \textbf{2nd Pass:} 
\STATE    \text{Run UCB} \cite{Auer et al.} \text{for the current sub-phase for time} $b_{2}$
\STATE    \text{Recommend the most played arm to the next sub-phase}
\STATE    \text{Keep doing it for the remaining time}
\end{algorithmic}
\end{algorithm}

\subsubsection{Description of \textbf{(pseudo) 2-pass-hybrid}}
For the set of $K$ arms $A$, arm memory $M\geq 2$, and total time horizon $T$, 
a. In the first pass we sample all arms for an equal time-horizon, the main idea is to get an estimate of empirically best arm, b. Recommend the empirically best arm to the second pass, and in the second pass run the similar allocation strategy that we've used in the previous algorithms, i.e. bifurcate the memory appropriately and recommend the most played arm to the next memory instance. The underlying idea being that once the best arm is being captured, it'll be further recommended with high probability. And since the the probability of making a mistake on the best arm is very low the expected regret will have reasonable bound order. \\
We find the necessary conditions for our \textbf{(pseudo)} 2-pass-hybrid to perform optimally and build \textbf{Algorithm-4} out of it.
 \\\\

\begin{lemma}
For given set $A$ of $K$ arms, $M$ arm memory, $\Delta_{\min}= \min_{\substack{i \in A}}\{\mu^{*}-\mu^{i}\}$, total time horizon $T$, then
\begin{itemize}
    \item The optimal total time duration($h_{0}b_{0}$) spent in 1st pass by \textbf{Algorithm-3} is,

\[
h_{0}b_{1}= \dfrac{h_{0}}{\Delta_{\min}^{2}}\log(1+\dfrac{\Delta_{\min}^{2}}{K}T^{2})
\]
where $h_{0}= \lceil \dfrac{K-1}{M-1} \rceil$. 

    \item And for the above optimal time duration spent in the first pass, the cumulative regret accrued ($R^{*}_{T}$) 
    
    \[
R_{T}^{*}= \dfrac{K}{\Delta_{\min}^{2}}\log(f(T))\times(1- \dfrac{1}{f(T)}) + \dfrac{T}{f(T)}
    \]
    where $f(T)= 1+\dfrac{\Delta_{\min}^{2}}{K}T^{2}$
\end{itemize}

\end{lemma}
\begin{proof}

 Let $X_{i}^{k}$ be the reward drawn from arm $i$ at time $k$, therefore $\overline{\mu_{i}}= \dfrac{\sum_{k=1}^{b_{0}} X^{k}_{i}}{b_{1}}$. We thus want a bound on $P_{i \neq i^{*}}(\overline{\mu_{i}} \geq \overline{\mu_{i^{*}}})$ or that sub-optimal arm is recommended. Let $E_{i}$ be the event such that 
 $\{\overline{\mu_{i}} \geq \overline{\mu_{i^{*}}}\}$
 
\begin{align*}
    P(E_{i})&= P(\overline{\mu_{i^{*}}}-\overline{\mu_{i}}- (\Delta_{i}) \leq -\Delta_{i} ) \\
     &= P(\dfrac{\sum{X^{j}_{i^{*}}}- \sum{X_{i}^{j}}}{b_{1}} - \Delta_{i} \leq -\Delta_{i}) \\
     &= P(\sum{X^{j}_{i^{*}}}- \sum{X_{i}^{j}} - b_{0}\Delta_{i} \leq -b_{0}\Delta_{i})
\end{align*}

Let $S_{2n}$= $\sum{X^{j}_{i^{*}}}- \sum{X_{i}^{j}}$, $t= -b_{1}\Delta_{i}$, so the above inequality can be rewritten as $P(S_{2n}- E[S_{2n}] \leq -t$, and thus using Hoeffding's inequality we get: 
 
\begin{align*}
     &P(S_{2n}- E[S_{2n}] \leq -t) \leq exp(\dfrac{-2t^{2}}{2b_{1}}) \\
     &P(S_{2n}- E[S_{2n}] \leq -t) \leq exp(\dfrac{-2(b_{1}^{2})(\Delta_{i}^{2})}{2b_{1}})\\
    \implies &P(S_{2n}- E[S_{2n}] \leq -t) \leq exp(-\Delta_{i}^{2}b_{1}) \\
    \implies &P(E_{i}) \leq \exp(-\Delta_{i}^{2}b_{1})
\end{align*}
 
From above we now know that $P(E_{i}) \leq exp(-\Delta_{i}^{2}b_{1})$. Now let $\overline{BA}$ be the event such that the recommended arm is \textbf{not} the best arm or P($\overline{BA}$)=1-P($BA$)= $1- P(\cap_{i \in [K]}\{\overline{E_{i}}\})$, where $E_{i}$ is the event that $ith$ arm is the recommended arm.  Thus we get,  
 
\begin{equation}
    1- P(\cap_{i \in [K]}\{\overline{E_{i}}\})= 1- \Pi_{i=1}^{K}(1-P(E_{i}))
\end{equation}
 
 Since we know that $P(E_{i}) \leq exp(-\Delta_{i}^{2}b_{1})$
 
 \begin{align*}
     P(\overline{BA}) &\leq 1- \Pi_{i=1}^{K}( 1- exp(-b_{1}\Delta_{i})) \\
     &\leq 1- \Pi_{i=1}^{K}(1- exp(-b_{1}\Delta_{\min})) \\
    &\leq 1-(1- K\times exp(-b_{1}\Delta_{\min})) \\
     &\leq K \times exp(-b_{1}\Delta_{\min})
 \end{align*}
 
 Thus, we get that the P(The best arm isn't recommended)= $P(\overline{BA}) \leq K \times exp(-b_{1}\Delta_{\min})$ or that the P(best arm is recommended)= $P(BA) \geq 1- K \times exp(-b_{1}\Delta_{\min})$
\\
\textbf{Calculating and optimizing the regret:}
 We'll calculate the total regret by bifurcating it in two passes. We know that for the first pass the regret accrues linearly for the worst case, or:
\begin{equation}
    R_{1} \leq Kb_{1}
\end{equation}

For the second pass however, it's a bit involved. Let $BA$ be the event that best arm is recommended by first pass at the beginning of the second pass, then we have:

\begin{equation}
    R_{2} \leq (1-P(BA))\times (h_{0}b_{2})+ P(BA)\times O(h_{0}\sqrt{\dfrac{\log(b_{2})}{b_{2}}})
\end{equation}
Which can be easily derived from using Lemma 4.4 in \cite{Chaudhari et al.} and arguing that in the worst case regret will be accrued linearly. For the sake of simplicity we ignore the second term in equation (3), and using the expression for $P(\overline{BA}) \leq K \times exp(-b_{1}\Delta_{\min})$ which we derived in the earlier section of the analysis, we convert our argument into a simple optimization problem where we want to find:

\begin{equation}
    b_{1, \text{opt}}, b_{2, \text{opt}}= \argmin_{b_{1}, b_{2}} (K\times b_{1}+e^{-b_{1}\Delta_{\min}^{2}}\times(h_{0}b_{1}))
\end{equation}
Such that,  $T= Kb_{1}+h_{0}b_{2}$.


Let $R= K\cdot b_{1}+e^{-b_{1}\Delta_{\min}^{2}}\cdot(h_{0}b_{2})$, substituting $b_{2}$ in terms of $b_{1}$ we get, 

\begin{equation}
    R= Kb_{1}+ e^{-b_{1}\Delta_{\min}^{2}}(T- Kb_{1})
\end{equation}

Differentiating wrt $b_{1}$ we get: 

\begin{align}
    \dfrac{dR}{db_{1}}= 0 \implies K + e^{-b_{1}\Delta_{\min}^{2}}(-K)+ \\ 
    \hspace{10pt}  \hspace{10pt} \hspace{10pt} \hspace{10pt}  e^{-b_{1}\Delta_{\min}^{2}}(T-Kb_{1})(-b_{1}\Delta^{2}_{\min})= 0 \\
    \implies K= e^{-b_{1}\Delta_{\min}^{2}}(K+b_{1}\Delta_{\min}^{2}(T-Kb_{1})) \\
    \implies e^{b_{1}\Delta_{i}^{2}}= \dfrac{K}{K+b_{1}\Delta_{\min}^{2}(T-Kb_{1})} \\
    \implies e^{b_{1}\Delta_{i}^{2}} \geq \dfrac{K}{K+b_{1}T\Delta_{\min}^{2}} \\
    \implies e^{b_{1}\Delta_{i}^{2}} \geq \dfrac{K}{K+\Delta_{\min}^{2}T^{2}} \hspace{10pt} \hspace{10pt} \hspace{10pt} \{ b_{1} < T\}\\
    \implies -b_{1}\Delta_{i}^{2} \geq \log(\dfrac{K}{K+\Delta^{2}_{\min}T^{2}}) \\ 
    \implies b_{1} \leq \dfrac{1}{\Delta_{\min}^{2}}\log(1+\dfrac{\Delta_{\min}^{2}}{K}T^{2})
\end{align}

Thus we get \textbf{(a).} 
\[
b_{1}= \dfrac{1}{\Delta_{\min}^{2}}\log(1+\dfrac{\Delta_{\min}^{2}}{K}T^{2})
\]
as the optimal time spent in the first pass. \\\\

Substituting the expression for $b_{1}= \dfrac{1}{\Delta_{\min}^{2}}\log(1+\dfrac{\Delta_{\min}^{2}}{K}T^{2})$ in  equation (6) and letting $f(T)= 1+\dfrac{\Delta_{\min}^{2}}{K}T^{2}$ we get, 
\textbf{(b).}
\[
R^{*}_{T}= \dfrac{K}{\Delta_{\min}^{2}}\log(f(T))\times(1- \dfrac{1}{f(T)}) + \dfrac{T}{f(T)}
\]

\end{proof}

Using \textbf{Lemma 5.4} we now propose a 2-pass \textbf{Algorithm-4}.

\begin{algorithm}[ht]
\caption{2-pass-hybrid}
\label{alg:algorithm_real2passhybrid}
\textbf{Input}: $A:$ the set of $K$ arms indexed by [K], $M(\geq 2)$: Arm memory size, $T:$ Total time horizon, $\Delta_{\min}= \min{i}\{\mu^{*}-\mu_{i}\}$ \\

\begin{algorithmic}[1] 

\IF {$M \geq K$}
\STATE Run UCB1 on $A$ until horizon
\ELSE
\STATE $\hat{a}= 1$. \{ Initial arm recommendation\}
\STATE $w= 1$. \{Counts the number of phases \}
\STATE $h_{0}= \lceil \dfrac{K-1}{M-1} \rceil$. \{ The number of sub-phases in a phase\}
\STATE $b_{1}= \dfrac{1}{\Delta_{\min}^{2}}\log(1+\dfrac{\Delta_{\min}^{2}}{K}T^{2})$ 
\STATE $b_{2}= T- b_{1}$ 
\WHILE{ the horizon is not finished}
\STATE $l= 0$
\FOR {$j= 1, \cdots, h_{0};$ if the horizon is not finished}
\STATE $S^{w,j}= \{l+1, \cdots, min\{l+1+(M-1), K\} \} $
\STATE $l=$ The highest arm index in $S^{w,j}$.
\IF { $\hat{a} \notin S^{w,j}$ }
\STATE $S^{w,j} = \{\hat{a}\} \cup S^{w,j}\setminus \{l\}$.
\STATE $l= l-1$
\ENDIF
\IF {w==1}
\STATE \{ALLOCATION STRATEGY\}
\STATE Play each arm in $S^{w,j}$ $b_{1}$ times.
\STATE \{RECOMMENDATION STRATEGY\}
\STATE $\hat{a}=$ empirically best arm. Or $\hat{a}= max_{i}\mu_{i}$
\ELSE
\STATE \{ALLOCATION STRATEGY\}
\STATE Run UCB1 on $S_{w,j}$ for horizon of $b_{w}$ pulls
or the remaining horizon; whichever is smaller.
\STATE \{RECOMMENDATION STRATEGY\}
\STATE $\hat{a}=$ The most played arm in $S^{w,j}$
\ENDIF
\ENDFOR
\STATE $w= w+1$. \{Increment phase count\}
\ENDWHILE
\ENDIF
\end{algorithmic}
\end{algorithm}

\section{Simulations}
We performed simulations to compare the performances of UCB-M from \cite{Chaudhari et al.}, UCB-LAM above, and standard UCB1. We set parameters to be K=30, M=4, and the bandit arms had means varying as $\mu_{i}= 0.99-0.1*i; i\in [0,K]$ with each arm $i$ following a Bernoulli distribution with mean $\mu_{i}$. The order of arrival was randomised and simulation shows the average over 10 simulations. 
\begin{figure}[H]
\centering
\includegraphics[width= 1\linewidth]{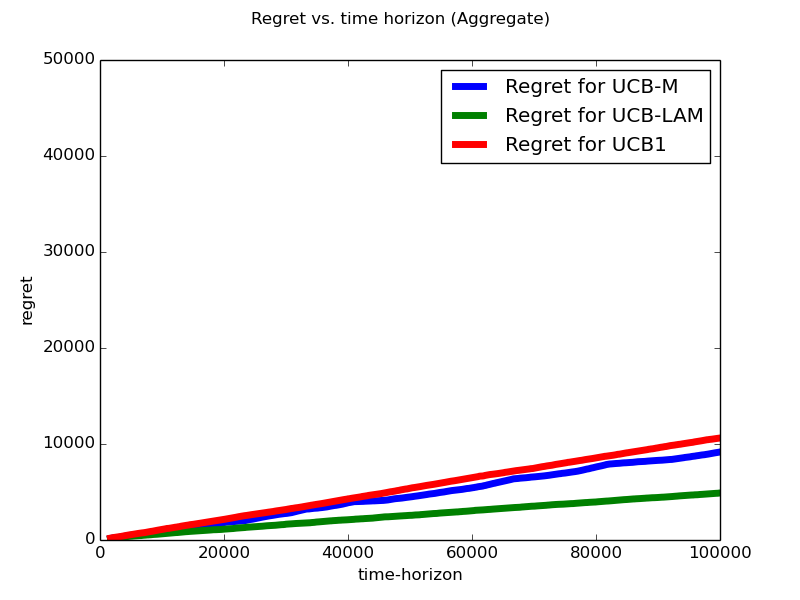}
\caption{Aggregate regret vs. Time horizon compared between UCB-M, UCB-LAM, UCB1}
\end{figure}

We thus observe that the order of regret is comparable across the algorithms, with UCB-LAM having $O(log(log(T)))$ number of passes over $O(log(T))$ number of passes of UCB-M.
\section{Conclusion}
We've thus explored the gap we set out to explore, which is the regret bound behaviour between one-pass \cite{Maiti et al.} and $log(T)$ passes. We've shown that in instance independent setting with no prior information about the system, $O(\sqrt{Tlog(T)})$ regret can be achieved with simply $log(log(T))$ passes instead of $log(T)$ passes. We've now als know that under large enough time horizons it's possible to achieve $O(\sqrt{Tlog(T)})$ regret with simply two passes. It makes sense intuitively because with the number of passes limited to two we want to learn the best arm behaviour as accurately as possible. However the behaviour of regret in smaller time horizons with constant number of passes needs to be explored further and we defer that to future studies. 

\section{Acknowledgement}
This work wouldn’t have been possible without timely feedback and discussions from Vishakha Patil (PhD student IISc Bangalore) and Dr. Arindam Khan (Assistant Professor, IISc Bangalore), especially with regards to setting the over-arching goal of exploring the domain between constant passes and $O(log(T))$ passes.
\\\\

\section{Appendix}

\appendix

\section{Proofs for Section 4}
\begin{lemma}
 For a given $K$- sized set of arms $A$, and an arm memory size $M < K$, the number of phases \textbf{Algorithm 1} executes is upper bounded by $x_{0}= 1+ \lceil \log(\log_{M(M+2)}( \dfrac{T}{h_{0}} )) \rceil$.
 \end{lemma}
\begin{proof}

 Firstly we'd like to make it clear that there exists\{ $K$,$M$ time horizon: $T$ \} such that the time horizon is complete while the algorithm is in between a particular phase. We have $b_{w}= (b_{w-1})^{2}$ or the length of each sub-phase in phase $w$ varies as $\{ b_{1}^{2^{1-1}}, b_{1}^{2^{2-1}}, b_{1}^{2^{3-1}}, \cdots, b_{1}^{2^{x_{0}-1}} \}$, where $x_{0}$ is the total number of phases or total number of passes. Now,

\begin{align}
T= \sum_{w= 1}^{x_{0}}\sum_{j= 1}^{h_{0}} b_{w,j} \\
= \sum_{w= 1}^{x_{0}} h_{0}b_{w} \cdots \cdots \{b_{w}= b_{w,1}= \cdots= b_{w, h_{0}}  \} \\
\Rightarrow T \geq h_{0}b_{x_{0}}    \\
\Rightarrow T \geq h_{0}(b_{1}^{(2^{x_{0}-1})}) \\
\Rightarrow \log_{b_{1}}(\dfrac{T}{h_{0}}) \geq 2^{x_{0}-1}\\
\Rightarrow \log\log_{b_{1}}(\dfrac{T}{h_{0}}) \geq x_{0} -1
\end{align}

Since $b_{1}= M(M+2)$
\begin{equation}
\Rightarrow x_{0} \leq \lceil \log\log_{M(M+2)}(\dfrac{T}{h_{0}}) \rceil +1
\end{equation}
Hence proved.
\end{proof}
\begin{lemma}
Consider events,

\begin{enumerate}

    \item A: Best arm(universal) is in the current ($w,j$) instance.
    \item B: Best arm(universal) is recommended to next instance (($w, j+1$) or ($w+1, 1$)).
    
\end{enumerate}
And let $J_{b_{w}}$ be the arm recommended by the recommendation strategy for \textbf{UCB-LAM}.
Then for a particular instance $s= (w,j)$, $M:$ arm-memory and $b_{w}:$ time-horizon, $P(B_{s}=1|A_{s}=1)= P(J_{b_{w}}= i^{*}) \geq 1-\dfrac{M-1}{(b_{w}/M -1)^{2}}$
 \end{lemma}
 \begin{proof}
 Using a side-result from \textbf{\textit{Lemma 1} in \cite{Bubeck et al.}} we get that whenever the most played arm $J_{b_{w}}$ is different from the optimal arm $i^{*}$ then at least one of the suboptimal arms $i$ is such that $T_{i}(b_{w}) \geq a_{i}b_{w}$, where $a_{1}, \cdots, a_{K}$ are real numbers such that $a_{1}+a_{2}+\cdots+a_{K}= 1$ and $a_{i} \geq 0,  \forall i$. \\
And that  $P\{T_{i}(b_{w}) \geq a_{i}b_{w}\} \leq \dfrac{1}{\alpha -1}(a_{i}b_{w} -1)^{2(1-\alpha)}$, where $T_{i}:$ number of pulls for arm $i$ and $\alpha$ refers to $UCB(\alpha)$ algorithm used during allocation strategy.\\\\ 

Now, we know that:

\begin{align*}
    P(J_{b_{w}}= i^{*})+ (\sum_{i \in A \setminus \{ i^{*}\}} P(J_{b_{w}}=i)) =1 \\
    \implies P(J_{b_{w}}= i^{*})= 1- \sum_{i \in A \setminus \{ i^{*}\}} P(J_{b_{w}}=i)
\end{align*}

\vspace{20pt}

When $i$ is the most played arm or $(J_{b_{w}}=i)$ we know that $T_{i} \geq b_{w}/K$ and from \textbf{Lemma A.5} which is derived from from \cite{Bubeck et al.} we know that

\begin{align*}
P_{i \neq i^{*}}(J_{b_{w}}= i) \leq \dfrac{1}{\alpha-1}(b_{w}/K-1)^{2(1-\alpha)}
\end{align*}

\begin{multline*}
     \implies 1- \sum_{i \in A \setminus \{ i^{*}\}} P(J_{b_{w}}=i) \geq \\ (1-\sum_{i \in A \setminus \{ i^{*}\}}\dfrac{1}{\alpha-1}(b_{w}/K-1)^{2(\alpha-1)})
\end{multline*}

\begin{align*}
    &\implies P(J_{b_{w}}= i^{*}) \geq 1- \dfrac{K-1}{\alpha -1}(b_{w}/K -1)^{2(\alpha-1)}
\end{align*}

Substituting arm memory: $M$, $\alpha= 2$ for our case, we get $P(J_{b_{w}}= i^{*}) \geq 1- \dfrac{M-1}{(b_{w}/M -1)^{2}}$\\\\
\end{proof}

\begin{corollary}
Consider events,

\begin{enumerate}

    \item A: Best arm(universal) is in the current ($w,j$) instance.
    \item B: Best arm(universal) is recommended to next instance (($w, j+1$) or ($w+1, 1$)).
    
\end{enumerate}
For an instance $s= (w,j)$ the probability of the optimal arm not being present in the current arm memory, $P(A_{s}= 0) \leq \dfrac{(M-1)\cdot h_{0}}{(b_{w-1}/M -1)^{2}} $
\end{corollary}
\begin{proof}
From \textbf{\textit{Lemma A.2}} above we know that (probability of recommending best-arm) \\

$P(B_{s}=1|A_{s}=1) \geq 1- \dfrac{M-1}{(b_{w}/M -1)^{2}}$.\\\\

Let $t_{w}= b_{w}/M$, and $m=(w-1, k^{'})$ be the memory instance in the previous phase $w-1$ containing the universal best arm $a^{*}$. \\\\

We know that $P(A_{s}= 1)= \Pi_{m=(w-1,k^{'})}^{(w,j-1)}P(B_{m}=1|A_{m}=1))$, because the probability that the current instance will have the best arm depends on the fact that previous instance had the best arm which it passed on -- all the way to the instance $(w-1, k)$ which actually had the best arm. Thus we get, 

\begin{multline*}
    \Pi_{m=(w-1,k^{'})}^{(w,j-1)}P(B_{m}=1|A_{m}=1))\geq\\\Pi_{m=(w-1,k^{'})}^{(w,j-1)} (1- \dfrac{M-1}{(t_{m}-1)^{2}})
\end{multline*}

We're considering the worst case scenario here when the best arm lies in the previous phase $w-1$. As going from $m= (w-1, k^{'})$ to $(w,j-1)$ has a phase change, $w-1$ to $w$, we bifurcate the expression in two terms, a. considering phase $w-1: k_{1}= h_{0}+1-k^{'}$ and considering phase $w: k_{2}= j-1$.

\begin{multline*}
     \implies \Pi_{m=(w-1,k^{'})}^{(w,j-1)}P(B_{m}=1|A_{m}=1))\geq \\ (1- \dfrac{M-1}{(t_{w-1} -1)^{2}})^{k_{1}}\times(1-\dfrac{M-1}{(t_{w}-1)^{2}})^{k_{2}} 
\end{multline*}


    
    

Where: $k_{1}+k_{2} \leq h_{0}$. Thus, 
\begin{multline*}
        1-\Pi_{m= (w-1, k^{'})}^{ (w, j-1)}P ( B_{m}= 1| A_{m}=1)) \leq \\ 1-(1- \dfrac{M-1}{(t_{w-1} -1)^{2}})^{k_{1}}\times(1-\dfrac{M-1}{(t_{w} -1)^{2}})^{k_{2}}
\end{multline*}

\begin{multline*}
    \implies P(A_{s}= 0) \leq \\ 1-(1-\dfrac{M-1}{(t_{w-1} -1)^{2}})^{k_{1}}\times(1-\dfrac{M-1}{(t_{w} -1)^{2}})^{k_{2}}
\end{multline*}

\begin{align*}
    \implies P(A_{s}= 0) \leq 1-(1- \dfrac{M-1}{(t_{w-1} -1)^{2}})^{h_{0}}
\end{align*}
    
    
    

The above expression is valid even in the case where the best arm lies in phase $w$ in instance $(w, k^{'})$, instead of $w-1$, with $j> k^{'}$. Primarily because $b_{w-1} < b_{w}$.\\

Using binomial expansion and $t_{w}= b_{w}/M$ thus we get, 

\[
P(A_{s}= 0) \leq \dfrac{(M-1)\cdot h_{0}}{(b_{w-1}/M -1)^{2}}
\]
\end{proof}
\vspace{30pt}
The significance of above lemmas is that they show, theoretically, that as we move ahead in time the probability that the best arm will be contained in the memory increases drastically which is what we guess intuitively.
\begin{lemma} 
Consider events, 
\begin{enumerate}

    \item A: Best arm(universal) is in the current ($w,j$) instance.
    \item B: Best arm(universal) is recommended to next instance (($w, j+1$) or ($w+1, 1$)).
    
\end{enumerate}
Let $s=(w,j)$ be the current instance with phase $w$ and sub-phase $j$, and let $r$ be the arm recommended to the next instance $(w, j+1) or (w+1, j)$ and let $\mu_{*}^{w,j}$ be the mean of that arm. Then $E[\mu^{*}-\mu_{*}^{w,j}]$ can be expressed in terms of $T_{1}, T_{2}$ or $b_{w}E[\mu^{*}-\mu_{*}^{w,j}]= T_{1}+T_{2}$, where: 
\begin{itemize}
    \item $T_{1}= P(A_{s}=1)\times \{b_{w}(E[\mu^{*}- \mu_{*}^{w,j}])P(B_{s}=0|A_{s}=1)$
    \item $T_{2}= P(A_{s}=0)\times \{b_{w}(E[\mu^{*}- \mu_{*}^{w,j}]) \}$
\end{itemize}
\end{lemma}

\begin{proof}
Consider given events: 
\begin{enumerate}

    \item A: Best arm(universal) is in the current ($w,j$) instance.
    \item B: Best arm(universal) is recommended to next instance (($w, j+1$) or ($w+1, 1$)).
    
\end{enumerate}
We know that:
\begin{itemize}
    \item $P(B=1 | A= 0)= 0$
    \item $P(B=0| A= 0)= 1$
    \item $P(B=1| A= 1)= x (\text{unknown})$
    \item $P(B=0| A= 1)= 1-x$
\end{itemize}
Now we know that $R^{1}_{w,j}$= $b_{w}E[\mu^{*}- \mu_{*}^{w,j}]$.  \\
Let $\textbf{s}:$ current memory instance ($w,j$). We know that
$E[E[X]]= E[X]$, thus we write, 
\\\\
\begin{align*}
    R^{1}_{w,j}&= P(A_{s}=1)\cdot \{ b_{w}(E[\mu^{*}- \mu_{*}^{w,j}])P(B_{s}=1|A_{s}=1)\}\\
   &+ P(A_{s}=1)\cdot \{b_{w}(E[\mu^{*}- \mu_{*}^{w,j}])P(B_{s}=0|A_{s}=1) \} \\
    &+ P(A_{s}=0)\cdot\{b_{w}(E[\mu^{*}- \mu_{*}^{w,j}])P(B_{s}=1|A_{s}=0) \} \\
    &+ P(A_{s}=0)\cdot \{b_{w}(E[\mu^{*}- \mu_{*}^{w,j}])P(B_{s}=0|A_{s}=0) \}
\end{align*}
Using $P(B|A)$ values above, we get,\\
\begin{align*}
    R^{1}_{w,j}&= P(A_{s}=1)\cdot \{b_{w}(E[\mu^{*}- \mu_{*}^{w,j}])P(B_{s}=0|A_{s}=1) \}\\
    &+ P(A_{s}=0)\cdot \{b_{w}(E[\mu^{*}- \mu_{*}^{w,j}]) \}
\end{align*}

Now from above, let: 

\begin{align*}
    T_{1}= P(A_{s}=1)\cdot \{b_{w}(E[\mu^{*}- \mu_{*}^{w,j}])P(B_{s}=0|A_{s}=1)\\
    T_{2}= P(A_{s}=0)\cdot \{b_{w}(E[\mu^{*}- \mu_{*}^{w,j}]) \}
\end{align*}
Hence, $R^{1}_{w,j}= b_{w}E[\mu^{*}-\mu_{*}^{w,j}]= T_{1}+T_{2}$
\end{proof}

\textbf{Lemma 4.1} \textit{Consider events, }
\begin{enumerate}

    \item \textit{A: Best arm(universal) is in the current ($w,j$) instance.}
    \item \textit{B: Best arm(universal) is recommended to next instance (($w, j+1$) or ($w+1, 1$)).}
    
\end{enumerate}
\textit{Let $s=(w,j)$ be the current instance with phase $w$ and sub-phase $j$, and let $r$ be the arm recommended to the next instance $(w, j+1) or (w+1, j)$ and let $\mu_{*}^{w,j}$ be the mean of that arm. Then if }
\begin{itemize}
    \item $T_{1}= P(A_{s}=1)\cdot \{b_{w}(E[\mu^{*}-\mu_{*}^{w,j}])P(B_{s}=0|A_{s}=1)$
    \item $T_{2}= P(A_{s}=0)\cdot \{b_{w}(E[\mu^{*}- \mu_{*}^{w,j}]) \}$
\end{itemize}
\textit{We have, \\}
$\dfrac{T_{1}+T_{2}}{b_{w}} \leq 2C\dfrac{(M-1)\cdot h_{0}}{(b_{w-1}/M -1)^{2}}(h_{0}+1)\sqrt{\dfrac{M\log(b_{w-1})}{b_{w-1}}}$ \\

\begin{proof}

We prove it in two steps, wherein first step we establish an upper-bound on $T_{1}$ and in the second step we establish an upper-bound on $T_{2}$. \\

\textbf{Step 1.} 
We know that $P(A_{s}=1) \leq 1$, and from \textbf{Lemma A.2}, $P(B_{s}=1|A_{s}=1)= P(J_{n}= i^{*}) \geq 1-\dfrac{M-1}{(b_{w}/M -1)^{2}} \implies P(B_{s}=0|A_{s}=1)= 1-P(J_{n}= i^{*}) \leq \dfrac{M-1}{(b_{w}/M -1)^{2}}$ \\

Thus using the probabilistic bounds from above and \textbf{Theorem 3.1} we get that, 

\begin{align*}
    T_{1} \leq 1\times \dfrac{M-1}{(b_{w}/M -1)^{2}} \times b_{w}C\sqrt{\dfrac{M\log(b_{w})}{b_{w}}}
\end{align*}

\textbf{Step 2.} Establishing bounds for $T_{2}$ isn't so straightforward and will thus be slightly more involved, since the best arm isn't present in the current instance $s=(w,j)$. Let $k$ be the minimum sub-phase in phase $w$ that has the best-arm $a^{*}$ or let $k= \min\{i\in[h_{0}]; a^{*}\in S^{w,i} \}$, we then analyse cases where a. $j \geq k$ and b. $j < k$ separately. let $\mu_{*}^{w,j}$ be the mean of that arm recommended by instance $(w,j)$ or $\mu_{*}^{w,j}$ 
\\
\textbf{Step 2.a} (for $j \geq k$) \\
We can write $E[\mu^{*}-\mu_{*}^{w,j}]= E[\mu^{*}-\mu_{*}^{w,k}]+\sum_{i=k}^{j-1}E[\mu_{*}^{w,i}-\mu_{*}^{w,i+1}]$ \\

From \textbf{Corollary 4.0.1} we know that approximate simple regret $E[r^{y}]= E[\mu_{*}^{w,i}-\mu_{*}^{w,i+1}]$ can be bounded by $C\sqrt{\dfrac{M\log(b_{w})}{b_{w}}}$. And using \textbf{Theorem 3.1} we can bound $E[\mu^{*}-\mu_{*}^{w,k}]$. We thus get, 

    
    
    


\begin{align*}
E[\mu^{*}-\mu_{*}^{w,j}] \leq Ch_{0}\sqrt{\dfrac{M\log(b_{w})}{b_{w}}}
\end{align*}

\textbf{Step 2.b}(for step $j< k$)\\

Let $l= \min\{i\in[h_{0}]; a^{*}\in S^{w-1,i} \}$, or the sub-phase in the previous phase containing best arm $a^{*}$. Thus we can then write,
\begin{align*}
   E[\mu^{*}-\mu_{*}^{w,j}] &\leq E[\mu^{*}-\mu_{*}^{w-1,l}] \\
  &+ \sum_{i=l}^{h_{0}-1}E[\mu^{w-1,i}_{*}-\mu_{*}^{w-1,i+1}] \\
  &+ E[\mu_{*}^{w-1,h_{0}}-\mu_{*}^{w,1}] \\
  &+ \sum_{i=1}^{j-1}E[\mu_{*}^{w,i}-\mu_{*}^{w,i+1}] 
\end{align*}


We'll now bound each of the above four terms, a. we know that $a^{*} \in S^{w-1,l}$ thus by \textbf{Theorem 3.1} we get that, $E[\mu^{*}- \mu_{*}^{w-1,l}] \leq C\sqrt{\dfrac{M\log(b_{w-1})}{b_{w-1}}}$, b. using \textbf{Corollary 4.0.1} we can bound $\sum_{i=l}^{h_{0}-1}E[\mu^{w-1,i}_{*}-\mu_{*}^{w-1,i+1}] \leq C(h_{0}-l)\sqrt{\dfrac{M\log(b_{w-1})}{b_{w-1}}}$ as it's nothing but the summation of consecutive recommended arms, c. for the last two terms similar to point (b) above, only now the time horizon for a sub-phase is $b_{w}$, we can again invoke \textbf{Corollary 4.0.1} and get the following, 


\begin{align*}
    E[\mu^{*}-\mu_{*}^{w,j}] &\leq C(h_{0}-l+1)\sqrt{\dfrac{M\log(b_{w-1})}{b_{w-1}}}\\
    &+Cj\sqrt{\dfrac{M\log(b_{w})}{b_{w}}}
\end{align*}

Since, $b_{w-1}<b_{w}$ \\ $\implies \sqrt{\dfrac{M\log(b_{w})}{b_{w}}}<\sqrt{\dfrac{M\log(b_{w-1})}{b_{w-1}}}$.
Thus we get that, 

\begin{align*}
    E[\mu^{*}-\mu_{*}^{w,j}] \leq C(h_{0}+j-l+1)\sqrt{\dfrac{M\log(b_{w-1})}{b_{w-1}}}
\end{align*}
Because there has to be atleast one instance containing best-arm $a^{*}$ in $h_{0}$ consecutive instances, $j \leq l$ and thus we get, 

\begin{align*}
    E[\mu^{*}-\mu_{*}^{w,j}] \leq C(h_{0}+1)\sqrt{\dfrac{M\log(b_{w-1})}{b_{w-1}}}
\end{align*}

Using results from \textbf{Step 2.a} and \textbf{Step 2.b} we get that $E[\mu^{*}-\mu_{*}^{w,j}] \leq C(h_{0}+1)\sqrt{\dfrac{M\log(b_{w-1})}{b_{w-1}}}$. Finally using \textbf{Corollary A.2.1} we get that, 

\begin{align*}
    T_{2} \leq \dfrac{(M-1)\cdot h_{0}}{(b_{w-1}/M -1)^{2}}b_{w}C(h_{0}+1)\sqrt{\dfrac{M\log(b_{w-1})}{b_{w-1}}}
\end{align*}

\textbf{Step 3} Combining results from \textbf{Step 1}, \textbf{Step 2}, and using the fact that $b_{w-1}<b_{w}$ along with \\

$\sqrt{\dfrac{M\log(b_{w})}{b_{w}}} < \sqrt{\dfrac{M\log(b_{w-1})}{b_{w-1}}}$ \\\\

we finally get that, 

\begin{align*}
    \dfrac{T_{1}+T_{2}}{b_{w}} \leq 2C\dfrac{(M-1)\cdot h_{0}}{(b_{w-1}/M -1)^{2}}(h_{0}+1)\sqrt{\dfrac{M\log(b_{w-1})}{b_{w-1}}}
\end{align*}
\end{proof}

\begin{lemma}
If a sequence $S$ is such that $S_{w}=(b)^{m^{w}/2}m^{w/2}$, then if 
$\sum(S)=\sum_{w=1}^{\log_{m}\log_{b}(T)} (b)^{m^{w}/2}m^{w/2}$, we have,
\begin{align*}
\sum(S) \leq \sqrt{\dfrac{m}{m-1}(\log_{b}(T))\cdot(T+ \dfrac{T^{1/m-1}}{b^{m/m-1}})}
\end{align*}
\end{lemma}
\begin{proof}
 We use \textit{\textbf{Cauchy-Schwartz Inequality}} which states that $\sum_{i=1}^{k}a_{i}b_{i} \leq \sqrt{\sum_{i=1}^{k}a^{2}_{i}}\cdot \sqrt{\sum_{i=1}^{k}b^{2}_{i}} $, where $a_{i}, b_{i} \in C, \forall i \in [1,k]$. 
\\\\

Let $K= \log_{m}\log_{b}(T)$, $A = \sqrt{\sum_{w=1}^{K} b^{m^{w}}}$, $B= \sqrt{\sum_{w=1}^{K}m^{w}}$, applying \textit{\textbf{Cauchy-Schwartz Inequality}} on $\sum(S)$ we get that, 

\begin{equation}
\sum(S) \leq A\cdot B
\end{equation}

\begin{align*}
B &= \sqrt{\sum_{w=1}^{K}m^{w}}= \sqrt{\dfrac{m(m^{K}-1)}{m-1}} \\
&= \sqrt{\dfrac{m}{m-1}(\log_{b}(T) - 1) }
\end{align*} 
\begin{equation}
\implies  B \leq (\sqrt{\dfrac{m}{m-1}(\log_{b}(T))})  
\end{equation}

Now, to bound $A$ consider, 
\begin{align*}
A^{2}&= \sum_{w=1}^{K} b^{m^{w}} \\   
A^{2}&= \sum_{w=1}^{\log_{m}\log_{b}(T)- 1}b^{m^{w}} + b^{m^{\log_{m}\log_{b}(T)}} \\
A^{2}&= \sum_{w=1}^{\log_{m}\log_{b}(T)- 1}b^{m^{w}} + T \\
\intertext{We know that, $\sum_{w=1}^{K-1} m^{w}= (m^{K}-m)/(m-1)$ }.
\intertext{Since $\sum a_{i} \leq \prod a_{i}, \forall a_{i} \geq 2$}
\intertext{Therefore, $\sum_{w=1}^{K-1} b^{m^{w}} \leq b^{\sum_{w=1}^{K-1}m^{w}}$}\\
\Rightarrow &\sum_{w=1}^{K-1} b^{m^{w}} \leq b^{(m^{K} - m)/(m - 1)} \\
\Rightarrow &\sum_{w=1}^{K-1} b^{m^{w}} \leq \dfrac{T^{1/m-1}}{b^{m/m-1}} \\
\Rightarrow &A^{2} \leq T+ \dfrac{T^{1/m-1}}{b^{m/m-1}}
\end{align*}

\begin{equation}
\Rightarrow A \leq (\sqrt{T+ \dfrac{T^{1/m-1}}{b^{m/m-1}}}) 
\end{equation}

Thus from (7),(8),(9) we get that: 
\[
\sum(S) \leq \sqrt{\dfrac{m}{m-1}(\log_{b}(T))*(T+ \dfrac{T^{1/m-1}}{b^{m/m-1}})}.
\]
\end{proof}

\textbf{Lemma 4.2} For $2 \leq M < K$, and for $T \geq KM^{2}(M+2)$, let $R^{(1)}= \sum_{w=1}^{x_{0}}\sum_{j=1}^{h_{0}} R^{(1)}_{w,j}$, then, 
\[ 
R^{1} \leq C_{0}+ C_{2} \times \log(\log_{b_{1}}(\dfrac{T}{h_{0}}))
\]
where $C_{2}$ is a constant depending on $K,M$.
\begin{proof}
We know that, 
\begin{align*}
R^{(1)}&= \sum_{w=1}^{x_{0}}\sum_{j=1}^{h_{0}}R^{(1)}_{w,j} \\
&= \sum_{j=1}^{h_{0}}R^{(1)}_{1,j}+ \sum_{w=2}^{x_{0}}\sum_{j=1}^{h_{0}}R^{(1)}_{w,j} \\
&\leq h_{0}b_{1} + \sum_{w=2}^{x_{0}}\sum_{j=1}^{h_{0}}R^{(1)}_{w,j} \\
&\leq h_{0}b_{1}+ \sum_{w=2}^{x_{0}}\sum_{j=1}^{h_{0}} b_{w}E[\mu^{*}- \mu^{w,j}_{*}]
\end{align*}

From \textbf{Lemma A.3} we know that $E[\mu^{*}- \mu^{w,j}_{*}]$ can be expressed in terms of $T_{1}, T_{2}$ and thus invoking \textbf{Lemma 4.1} we get,

\begin{align*}
R^{1} & \leq h_{0}b_{1}\\ 
&+\sum_{w=2}^{x_{0}}h_{0}b_{w}(2C\dfrac{(M-1)\cdot h_{0}(h_{0}+1)}{(b_{w-1}/M -1)^{2}}\sqrt{\dfrac{M\log(b_{w-1})}{b_{w-1}}})\\
\intertext{Let $C_{1}= 2C\dfrac{(M-1)\cdot h_{0}^{2}(h_{0}+1)}{1}\sqrt{M}$, then}\\
R^{1} &\leq h_{0}b_{1}+C_{1}\sum_{w=2}^{x_{0}}\dfrac{b_{w}}{(b_{w-1}/M-1)^{2}}\sqrt{\dfrac{\log(b_{w-1})}{b_{w-1}}}\\
\intertext{Using the fact that $\dfrac{\log(x)}{x} \leq 1/2,  \forall x > 1$ we get,}\\  
R^{1} &\leq h_{0}b_{1}+\dfrac{C_{1}}{\sqrt{2}}\sum_{w=2}^{x_{0}} \dfrac{b_{w}}{(b_{w-1}/M -1)^{2}}\\
\intertext{Since $b_{w}=b_{w-1}^{2}$ we can write,} \\
R^{1} &\leq h_{0}b_{1}+\dfrac{C_{1}M^{2}}{\sqrt{2}}\sum_{w=2}^{x_{0}} \dfrac{1}{(1-\dfrac{M}{b_{w-1}})^{2}} \\
 &\leq h_{0}b_{1}+ \dfrac{C_{1}M^{2}}{\sqrt{2}}\sum_{w=2}^{x_{0}} \dfrac{1}{(1-\dfrac{M}{b_{1}})^{2}}\\
&\leq h_{0}b_{1}+ \dfrac{C_{1}M^{2}}{\sqrt{2}(1-\dfrac{M}{b_{1}})^{2}}\times (\log(\log_{b_{1}}(\dfrac{T}{h_{0}})))
\end{align*}

Thus we get $ R^{1} \leq  C_{0}+ C_{2} \times \log(\log_{b_{1}}(\dfrac{T}{h_{0}}))$, where $C_{2}= \dfrac{C_{1}M^{2}}{\sqrt{2}(1-\dfrac{M}{b_{1}})^{2}}$ is a constant depending on $K, M$.
\end{proof}
\textbf{Lemma 4.4}  \textit{For $2 \leq M < K$, and for $T \geq KM^{2}(M+2)$, let,  $R^{(2)}= \sum_{w=1}^{x_{0}}\sum_{j=1}^{h_{0}} R^{(2)}_{w,j}$, then,}  

\[
R^{(2)} \leq C_{0}+ C_{4}\sqrt{(\log_{b_{1}}(\dfrac{T}{h_{0}}))*(\dfrac{T}{h_{0}})}
\]
\textit{Where $C_{0} , C_{4}$ are constants depending on K and M.}
\begin{proof}
 For any sub-phase $j$ of any phase $w \geq 2$, due to Lemma 3, we know that there exists a constant $C$ such that $R^{(2)}_{w,j} \leq C \sqrt{b_{w}M\log(b_{w})}$. Therefore,

\begin{align*}
R^{(2)}&= \sum_{w=1}^{x_{0}}\sum_{j=1}^{h_{0}}R^{(2)}_{w,j} \\
&= \sum_{j=1}^{h_{0}}R^{(2)}_{1,j}+ \sum_{w=2}^{x_{0}}\sum_{j=1}^{h_{0}}R^{(2)}_{w,j}
\end{align*}

Now $R^{(2)}_{w,j} = \sum_{t=1}^{b_{w}} (E[\mu_{*}^{w,j}] - E[\mu_{a_{t}}])$, and we know that $\mu_{*}^{w,j} \leq \max_{a \in S^{w,j}}\mu_{a}$. Thus using \textbf{Lemma 4.3} we get that, 

\begin{align*}
R^{(2)} & \leq h_{0}b_{1}+ C\sum_{w=2}^{x_{0}}\sum_{j=1}^{h_{0}} \sqrt{b_{w}M\log(b_{w})} \\
&\leq h_{0}b_{1}+ Ch_{0}\sum_{w=2}^{x_{0}} \sqrt{(b_{1})^{2^{w-1}}2^{w-1} (M\log(b_{1}))} \\
&\leq h_{0}b_{1}+ Ch_{0}(M\log(b_{1}))\sum_{w=1}^{x_{0}-1} \sqrt{(b_{1})^{2^{w}}2^{w}} \\
&\leq h_{0}b_{1}+ Ch_{0}(M\log(b_{1}))\sum_{w=1}^{\log\log_{b1}(T/h_{0})} \sqrt{(b_{1})^{2^{w}}2^{w}} \\
\end{align*}

Let $C_{3}= C(\dfrac{K-1}{M-1})M\log(M(M+2))$, \\
and $C_{0}=  \dfrac{(K-1)(M)(M+2)}{(M-1)}$. \\

\begin{align*}
&\leq h_{0}b_{1}+ C_{3}\sum_{w=1}^{\log\log_{b1}(T/h_{0})} \sqrt{(b_{1})^{2^{w}}2^{w}} \\
\intertext{Invoking \textbf{Lemma A.4} and using $m=2$ we get} \\
&\leq h_{0}b_{1}+ C_{3}\sqrt{(\log_{b_{1}}(T/h_{0}))\cdot(T/h_{0}+ \dfrac{(T/h_{0})^{1/(1)}}{b_{1}^{2/(2-1)}})}\\
\intertext{Thus we get,} \\
R^{(2)} &\leq C_{0}+ C_{4}\sqrt{(\log_{b_{1}}(\dfrac{T}{h_{0}}))*(\dfrac{T}{h_{0}})}\\\\
\intertext{where $C_{4}= C(\dfrac{K-1}{M-1})M\log(M(M+2))\sqrt{1+\dfrac{1}{b_{1}}}$} \\
\intertext{and $C_{0}=  \dfrac{(K-1)(M)(M+2)}{(M-1)}$} \\
\end{align*}

\end{proof}

\vspace{50pt}

\begin{lemma}[From Lemma 1 in \cite{Bubeck et al.}]
Let $a_{1}, \cdots, a_{K}$ be real numbers such that $a_{1}+\cdots+a_{K}=1$ and 
$a_{i} \geq 0$ for all $i$, with the additional property that for all suboptimal arms $i$ and all optimal arms $i^{*}$, one has $a_{i} \leq a_{i^{*}}$. Let $n$: total time horizon, and $J_{n}$: max played arm recommended.Then for $\alpha > 1$, and sufficiently large time horizons $a_{i}n \geq 1+\dfrac{4\alpha ln(n)}{\Delta_{i}^{2}}$, the allocation strategy given by UCB($\alpha$) associated with the recommendation given by the most played arm ensures that the probability of the suboptimal arm being the max played arm:

\begin{align*}
    P_{i \neq i^{*}}(J_{n}= i) \leq \dfrac{1}{\alpha-1}(n/K-1)^{2(1-\alpha)}
\end{align*}

\end{lemma}

\begin{proof}

We first prove that whenever the most played arm $J_{n}$ is different from an optimal arm $i^{*}$, then at least one of the suboptimal arms i is such that $T_{i}(n) \geq a_{i}n$. To do so, we use a contrapositive method and assume that $T_{i}(n) < a_{i}n$ for all suboptimal arms. Then, 

\begin{align*}
    (\sum_{i=1}^{K}a_{i})n= n = \sum_{i=1}^{K}T_{i}(n) < \sum_{i_{*}}T_{i_{*}}(n)+ \sum_{i}a_{i}n
\end{align*}

where, in the inequality, the first summation is over the optimal arms, the second one, over the suboptimal ones. Therefore, we get

\begin{align*}
    \sum_{i^{*}}a_{i^{*}}n < \sum_{i^{*}}T_{i^{*}}(n)
\end{align*}

and there exists at least one optmial arm $i^{*}$ such that $T_{i^{*}} > a_{i^{*}}n$. Since by definition of the vector $(a_{1},\cdots,a_{K})$, one has $a_{i} \leq a_{i^{*}}$ for all suboptimal arms, it comes that $T_{i} < a_{i}n \leq a_{i^{*}}n < T_{i^{*}}(n)$ for all suboptimal arms, and the most played arm $J_{n}$ is thus an optimal arm. \\\\

A side-result extracted from [\cite{Audibert et al.}, proof of Theorem 7], see also [\cite{Auer et al.}, proof of Theorem 1], states that for all suboptimal arms $i$ and rounds $t \geq K+1$, 

\begin{align*}
    P\{I_{t}=i, T_{i}(t-1) \geq l\} \leq 2t^{1-2\alpha} \hspace{30pt} (l\geq \dfrac{4\alpha ln(n)}{\Delta_{i}^{2}})
\end{align*}

We denote by $\lceil x \rceil$ the upper integer part of a real number $x$. For a suboptimal arm $i$ and since by the assumptions on $n$ and $a_{i}$, the choice $l= \lceil a_{i}n \rceil-1$ satisfies $l \geq K+1$ and $l \geq (4\alpha ln(n))/\Delta_{i}^{2}$, 

\begin{align*}
    P\{T_{i}(n) \geq a_{i}n\} &= P\{T_{i}(n) \geq \lceil a_{i}n \rceil\} \\
    &\leq \sum_{t= \lceil a_{i}n \rceil}^{n}P\{T_{i}(t-1)= \lceil a_{i}n \rceil-1, I_{t}=i\}\\
    &\leq \sum_{t= \lceil a_{i}n \rceil}^{n}2t^{1-2\alpha} \leq 2\int_{\lceil a_{i}n \rceil -1}^{\infty} v^{1-2\alpha} \,dv \\
    &\leq \dfrac{1}{\alpha-1}(a_{i}n-1)^{2(1-\alpha)}, 
\end{align*}
With the uniform choice of $a_{i}= 1/K$ we finally get, 

\begin{align*}
    P_{i \neq i^{*}}(J_{n}= i) \leq P(T_{i}(n) \geq n/K) \leq \dfrac{1}{\alpha-1}(n/K-1)^{2(1-\alpha)}
\end{align*}
\end{proof}

\section{Proofs for Section 5}
\begin{lemma}
Total number of passes in 2-pass UCB-LAM is two.
\end{lemma} 

\begin{proof}
 From Algorithm-2 above we know that $b_{1}= \dfrac{\sqrt{1+4\dfrac{T}{h_{0}}}-1}{2}$ and 
$b_{2}= \dfrac{2\dfrac{T}{h_{0}}+1-\sqrt{1+\dfrac{4T}{h_{0}}}}{2}$. So the total time taken for the first two passes($T_{\text{two-pass}}$) is:

\begin{align*}
    T_{\text{two-pass}} &= h_{0}\times b_{1}+ h_{0}\times b_{2} \\
     &= \dfrac{h_{0}\sqrt{1+4\dfrac{T}{h_{0}}}-1}{2} + \dfrac{2T+h_{0}-h_{0}\sqrt{1+\dfrac{4T}{h_{0}}}}{2} \\
    & \implies T_{\text{two-pass}}= T\\
\end{align*}

Thus we see that the total time horizon is depleted by the end of second pass, and thus the total number of passes in Algorithm-2 in two.
\end{proof}

\textbf{Lemma 5.1} \textit{For $2 \leq M < K$, and total time-horizon $T$, \\ we get that, for Algorithm-2,}
\[
R_{2} \leq C_{2}+C_{0}\sqrt{T+0.25h_{0}}+C_{1}\sqrt{T\log(T/h_{0})}
\]
\textit{Where $C_{0}, C_{1}, C_{2}$ are constants depending on $M,K$.}
\begin{proof}
 We know that $R_{2}$ can be written as, 

\begin{align*}
    R_{2}&= \sum_{w=1}^{2}\sum_{j=1}^{h_{0}}(R^{(2)}_{w,j}) \\
    &= \sum_{w=1}^{2}\sum_{j=1}^{h_{0}}(\sum_{t=1}^{b_{w}} (E[\mu_{*}^{w,j}] - E[\mu_{a_{t}}])) \\ 
    &\leq h_{0}b_{1}+ \sum_{j=1}^{h_{0}}(\sum_{t=1}^{b_{2}}(E[\mu_{*}^{2,j}] - E[\mu_{a_{t}}])) \\
\end{align*}

From \textbf{Lemma 4.3} we get that given a set of $K$ arms as the input for any horizon $T$, the cumulative regret incurred by $UCB1$ when $T \geq K/2$, is upper-bounded by $18\sqrt{TK\log(T)}$ or $R_{T}^{*} \leq 18\sqrt{TK\\log(T)}$. And we know that $\mu_{*}^{w,j} \leq \max_{a \in S^{w,j}} \mu_{a} \implies E[\mu^{w,j}_{*}-\mu_{a_{t}}] \leq E[\max_{a\in S^{w,j}}\mu_{a}- \mu_{a_{t}}]$. Thus, 

\begin{align*}
    R_{2} \leq h_{0}b_{1}+ h_{0}C\sqrt{b_{2}M\log(b_{2})}
\end{align*}

Substituting $b_{1}= \dfrac{\sqrt{1+4\dfrac{T}{h_{0}}}-1}{2}$ and 
$b_{2}= \dfrac{2\dfrac{T}{h_{0}}+1-\sqrt{1+\dfrac{4T}{h_{0}}}}{2}$, we get, 

\begin{align*}
    R_{2} \leq 0.5(\sqrt{h_{0}^{2}+4h_{0}T}-h_{0})+C_{1}\sqrt{T\log(T/h_{0})}
\end{align*}
where $C_{1}=f(M,K), C_{0}= g(M,K), C_{2}= h(M, K)$\\ 
Thus we get that 

\[
R_{2} \leq C_{2}+C_{0}\sqrt{T+0.25h_{0}}+C_{1}\sqrt{T\log(T/h_{0})}.
\]
\end{proof}

\textbf{Lemma 5.2} \textit{For $2 \leq M < K$, and total time-horizon $T$, \\ we get that, for Algorithm-2,\\  $R_{1}= \sum_{w=1}^{2}\sum_{j=1}^{h_{0}}(R^{(1)}_{w,j}) \leq O(\sqrt{T})$}

\begin{proof}
We know that $R_{1}$ can be written as,\\ 

 $R_{1}$= $\sum_{w=1}^{2}\sum_{j=1}^{h_{0}}\sum_{t=1}^{b_{w}}(E[\mu^{*}-\mu_{*}^{w,j}])$
 
\begin{align*}
\implies &R_{1} = \sum_{w=1}^{2}\sum_{j=1}^{h_{0}}b_{w}(E[\mu^{*}-\mu_{*}^{w,j}]) \\
& R_{1} \leq h_{0}b_{1}+ \sum_{j=1}^{h_{0}}b_{2}(E[\mu^{*}-\mu_{*}^{2,j}])
\end{align*}

From \textbf{Lemma A.3} we know that $b_{2}E[\mu^{*}-\mu_{*}^{2,j}]$ can be written in terms of $T_{1}$ and $T_{2}$, or $b_{2}E[\mu^{*}-\mu_{*}^{2,j}]= T_{1}+T_{2}$, where, 

\begin{align*}
    T_{1}&= P(A_{s}=1)\times \{b_{w}(E[\mu^{*}- \mu_{*}^{w,j}])P(B_{s}=0|A_{s}=1)\\
    T_{2}&= P(A_{s}=0)\times \{b_{w}(E[\mu^{*}- \mu_{*}^{w,j}]) \}
\end{align*}

Where $s=(w,j)$ the current instance and $\mu_{*}^{w,j}$ is the mean of the arm recommended to the next instance. And invoking \textbf{Lemma 4.1} will give us that, 

\begin{align*}
 T_{1}+T_{2} \leq 2Cb_{2}\dfrac{(M-1)\cdot h_{0}}{(b_{1}/M -1)^{2}}(h_{0}+1)\sqrt{\dfrac{M\log(b_{1})}{b_{1}}}
\end{align*}
Thus we get that,

\begin{align*}
    R_{1}& \leq h_{0}b_{1}+2Cb_{2}\dfrac{(M-1)\cdot h_{0}^{2}}{(b_{1}/M -1)^{2}}(h_{0}+1)\sqrt{\dfrac{M\log(b_{1})}{b_{1}}} \\
    &\text{Substituting values for} b_{1}, b_{2} for \textbf{Algorithm-2} \\
    b_{1}&= \dfrac{\sqrt{1+4\dfrac{T}{h_{0}}}-1}{2}\\
    b_{2}&= \dfrac{2\dfrac{T}{h_{0}}+1-\sqrt{1+\dfrac{4T}{h_{0}}}}{2}= b_{1}^{2}
\end{align*}
in the above inequality we finally get that $R_{1} \leq O(\sqrt{T})$.
\end{proof}

\bibliographystyle{named}


\end{document}